\documentclass{article}

 \usepackage[preprint]{neurips_2025}

\usepackage[utf8]{inputenc} 
\usepackage[T1]{fontenc}    
\usepackage{hyperref}       
\usepackage{url}            
\usepackage{booktabs}       
\usepackage{amsfonts}       
\usepackage{nicefrac}       
\usepackage{microtype}      
\usepackage{xcolor}         
\usepackage{graphicx}
\usepackage{caption}
\usepackage{wrapfig, blindtext} 
\usepackage{makecell}
\usepackage{adjustbox}
\usepackage{amsmath}
\newcommand{\ie}{\emph{i.e.}}

\newcommand{\etc}{\emph{etc.}}
\renewcommand{\vec}[1]{\boldsymbol{#1}}    
\newcommand{\Q}{\vec{Q}}
\newcommand{\K}{\vec{K}}
\newcommand{\V}{\vec{V}}
\usepackage{hyperref}
\usepackage{tcolorbox} 
\title{DreamVE: Unified Instruction-based Image and Video Editing}

\author{	Bin Xia $^{1}$, Jiyang Liu  $^{2}$, Yuechen Zhang$^{1}$, Bohao Peng $^{1}$, Ruihang Chu$^{1}$, \\\textbf{Yitong Wang}$^{2}$, \textbf{Xinglong Wu}$^{2}$, \textbf{Bei Yu}$^{1}$, and \textbf{Jiaya Jia}$^3$ \\
	$^{1}$ CUHK, $^2$  ByteDance Inc, $^3$ HKUST  \\ \href{https://zj-binxia.github.io/DreamVE-ProjectPage/}{https://zj-binxia.github.io/DreamVE-ProjectPage/}
}

\begin{document}

\maketitle

\begin{center}
    \vspace{-5mm}
    \captionsetup{type=figure}
    \includegraphics[width=.928\linewidth]{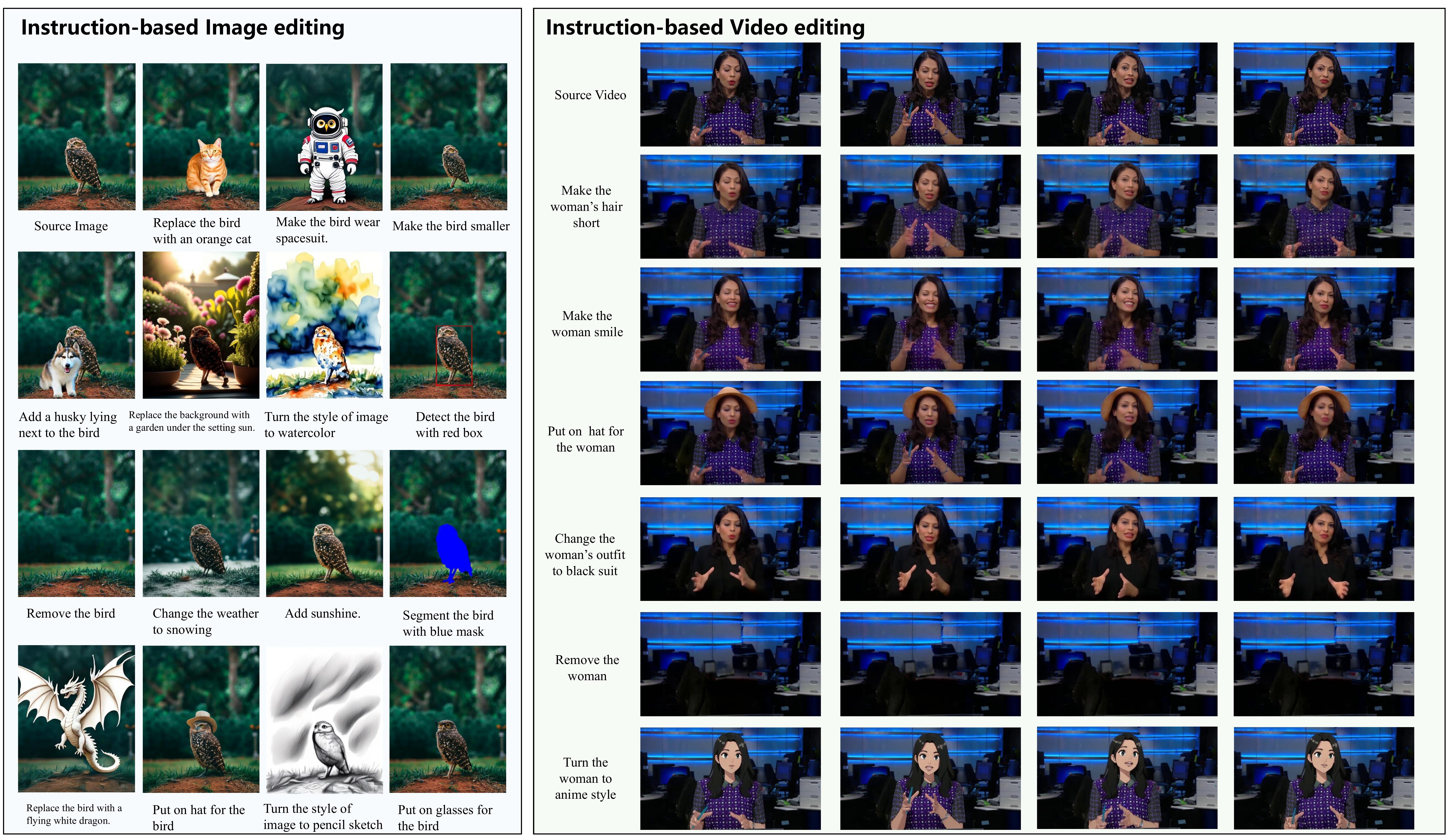}
    \vspace{-1mm}
    \captionof{figure}{The gallery of DreamVE on instruction-based image and video editing tasks. }
    \label{fig:summary}
\end{center}

\begin{abstract}
Instruction-based editing holds vast potential due to its simple and efficient interactive editing format. 
However, instruction-based editing, particularly for video, has been constrained by limited training data, hindering its practical application.
To this end, we introduce DreamVE, a unified model for instruction-based image and video editing. Specifically, We propose a two-stage training strategy: first image editing, then video editing. This offers two main benefits: (1) Image data scales more easily, and models are more efficient to train, providing useful priors for faster and better video editing training. (2) Unifying image and video generation is natural and aligns with current trends. Moreover, we present comprehensive training data synthesis pipelines, including collage-based and generative model-based data synthesis. 
The collage-based data synthesis combines foreground objects and backgrounds to generate diverse editing data, such as object manipulation, background changes, and text modifications. It can easily generate billions of accurate, consistent, realistic, and diverse editing pairs. We pretrain DreamVE on extensive collage-based data to achieve strong performance in key editing types and enhance generalization and transfer capabilities. However, collage-based data lacks some attribute editing cases, leading to a relative drop in performance. In contrast, the generative model-based pipeline, despite being hard to scale up, offers flexibility in handling attribute editing cases. Therefore, we use generative model-based data to further fine-tune DreamVE. Besides, we design an efficient and powerful editing framework for DreamVE. We build on the SOTA T2V model and use a token concatenation with early drop approach to inject source image guidance, ensuring strong consistency and editability. The codes and models will be released.
\end{abstract}

\vspace{-4mm}
\section{Introduction}
\label{sec:intro}
\vspace{-1mm}

Recently, text-to-video (T2V) generative foundation models~\cite{sora,kling,gen3} have made significant advancements in creating highly stable and realistic videos. Moreover, video editing focuses on modifying source videos to meet user requirements, unlocking a wide range of potential applications. The powerful T2V models further enable video editing to reach the performance required for applications.



Video editing can be categorized into training-free and training-based 
methods.  Most current research focuses on training-free editing due to the lack of video editing data and the high costs of training. However, training-free editing faces two main issues: \textbf{(1)} relatively low success rates, requiring complex parameter adjustments, which is user-unfriendly; \textbf{(2)} challenges in continuous improvement and optimization. \textbf{(3)} the inversion process requires additional computation, which is inefficient. Thus, training-based editing is key to practical applications. Among these, instruction-based editing stands out, allowing users to modify videos or images with text commands, being more user-friendly.



The main challenge in instruction-based editing is the lack of high-quality, accurate training data. Although some methods~\cite{instructp2p,instructv2v} use the prompt-to-prompt method~\cite{prompt2prompt} to generate paired image or video editing data, they face several issues: \textbf{(1)} They typically use classic Unet-based models~\cite{LDM} for synthesis, leading to poor synthetic quality.   \textbf{(2)} The acceptance rate for synthesizing accurate editing data is very low, typically lower $10\%$. \textbf{(3)} Filtering successful edits is challenging, even when leveraging GPT-4o for selection, as hallucination issues lead to incorrect filtering. \textbf{(4)} Running large generative models is slow and resource-intensive. These issues make scaling data difficult, limiting the generalization and performance of instruction-based editing.


To this end, we propose DreamVE, a unified instruction-based image and video editing model, with focus on training strategy, data construction, and framework design. For the training scheme, we first train on image editing data, then fine-tune on video editing. Image editing priors accelerate video training and improve performance. This scheme offers two advantages: \textbf{(1)} Image data scales more easily and trains faster, reducing reliance on video data and compute. \textbf{(2)} The recent success of GPT-4o in unified image generation and editing demonstrates the growing trend toward unification, and extending this paradigm to both image and video editing aligns well with future research directions.

\begin{figure*}[t]
	\centering
 \resizebox{1\linewidth}{!}{
	\includegraphics[height=8cm]{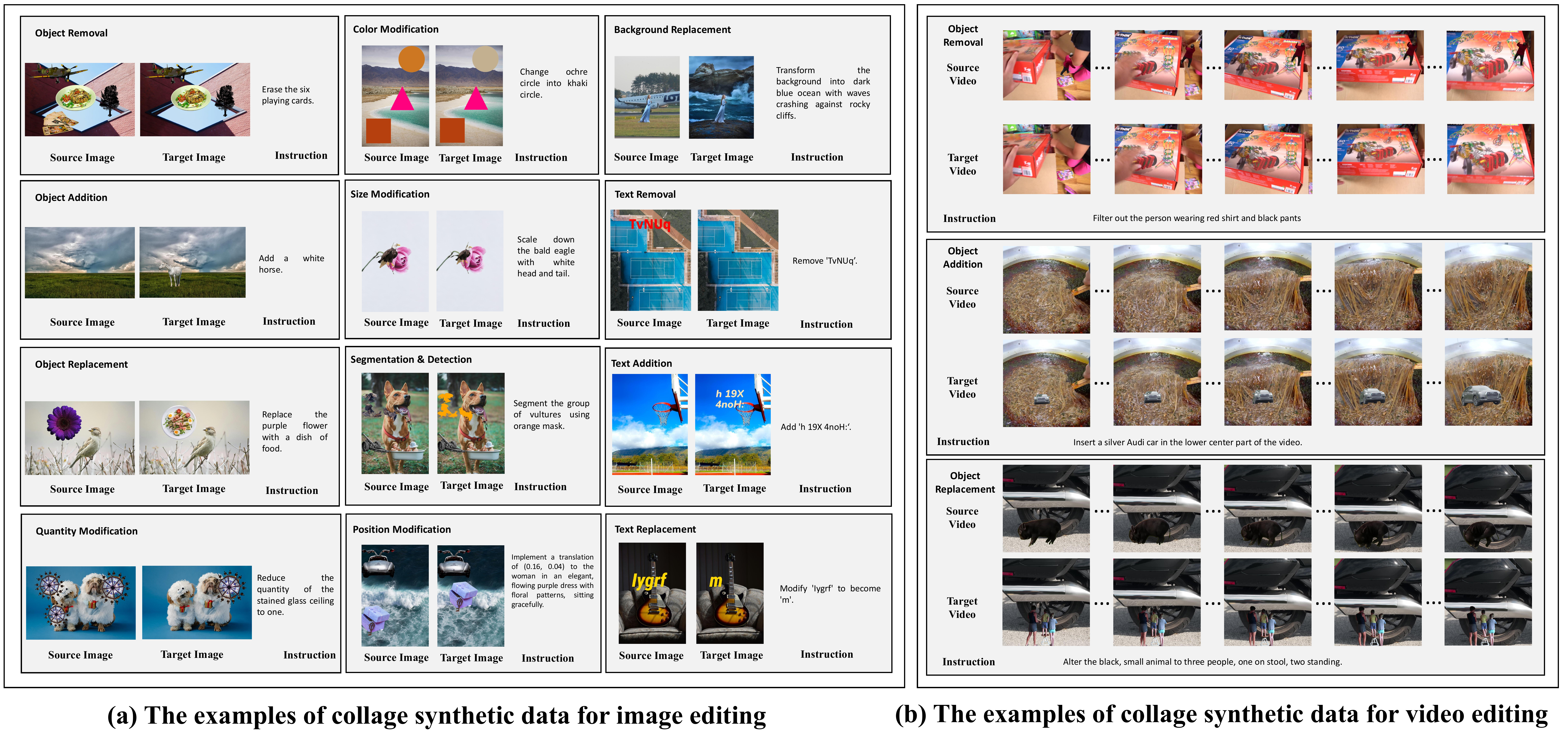}
 }
  \vspace{-5mm}
	\caption{ The example of collage-based synthetic data. 
\textbf{(a)} Example of instruction-based image editing data, including operations like object removal, addition, replacement, quantity/color/size/position modification, background replacement, and text editing (removal, addition, replacement).
\textbf{(b)} Example of instruction-based video editing data, including object removal, addition, and replacement.
}
\label{fig:collageSyn}
\vspace{-4mm}

\end{figure*}

 For image data construction, DreamOmni~\cite{dreamomni} utilized collage-based pipeline to scale up image editing training data on object addition, removal, and replacement, yielding promising results.
We further explore the potential of collage-based data by incorporating additional editing tasks, such as background replacement and modifications to object quantity, size, color, position, and text (Fig.~\ref{fig:collageSyn}~(a)). This approach enabled us to easily construct a $5M$ image dataset for DreamVE's pretraining.
 For video data, we specifically created a video foreground dataset using segmentation service from open-source video datasets, and used our video collage synthetic data pipeline to generate extensive video editing data for tasks like object addition, deletion, and replacement.

Although synthetic collage data has achieved excellent editing performance, we observed a drop in response for some corner attribute editing cases, such as modifying actions, expressions, \etc. To this end, we design a generative model-based synthetic data pipeline.  Instead of using prompt-to-prompt~\cite{prompt2prompt} to make editing data with SD1.5~\cite{instructp2p,instructv2v}, we leveraged state-of-the-art (SOTA) DIT-based models to make image and video editing data. However, since recent DIT models~\cite{flux,hunyuanvideo} use Flow Matching sampling~\cite{flowmatching} and distilled CFG, prompt-to-prompt was not directly applicable. To overcome this, we propose a feature-mixing approach.  Scaling image and video editing data is challenging. Thanks to extensive collage-based editing data pretraining, the DreamVE requires only minimal fine-tuning data to adapt to corner editing cases, showing powerful generalization.


For editing framework, we use DIT-based T2V model as the base and explore two ways to inject the source image as guidance into the T2V network: \textbf{(1)} concatenating the noisy latent and source image features along the channel dimension (similar to InstructP2P in Fig.~\ref{fig:framework}~(a)); \textbf{(2)} our concatenating the noisy latent and source image features along the token dimension with early drop (Fig.~\ref{fig:framework}~(b)). The ``early drop'' means the source image features only need to be processed in the first four DIT blocks, after which they can be discarded in the following layers.
The second method significantly outperforms the first under the same training conditions, with both methods having nearly identical runtime. Notably, our token concatenation scheme is not only efficient but also highly scalable, making it well-suited for future unified model research~\cite{dreamomni,omnigen}.




\begin{itemize}

\item We introduce DreamVE, a unified instruction-based editing model trained in two stages: image editing followed by video editing. This approach reduces video training costs, improves performance, and enables unified image and video editing. DreamVE significantly outperforms existing open-source models on both tasks.

\item We propose a comprehensive synthetic collage data pipeline for instruction-based image and video editing. Collage-based data is accurate, realistic, diverse, and scalable, covering key editing tasks. Pretraining DreamVE on this data achieves strong performance.

\item We propose a generative model-based synthetic data pipeline. Compared to previous methods, we use SOTA DIT-based models to generate higher-quality data. Besides, we introduce a feature-mixing scheme to adapt to these models for data generation.

\item For the editing framework, we explore various image conditioning injection schemes and adopt the optimal token concatenation with early drop scheme in DreamVE.

\end{itemize}

\vspace{-4mm}
\section{Related Work}
\vspace{-1mm}

\noindent\textbf{Image and Video Synthesis.} 
Diffusion models~\cite{sohl2015deep,DDPM,DDPM3,DDPM4,DDPM5,DDPM6,batzolis2021conditional,song2020score,glide} have achieved great success in text-to-image (T2I) generation. The latent diffusion model~\cite{LDM} applies the diffusion process in a VAE ~\cite{VAE} compressed latent space and uses cross-attention to guide image generation with text. This has greatly enhanced the usability of image synthesis models, leading to their widespread adoption. Subsequently, an increasing number of T2I models have been developed, such as Imagen~\cite{imagen}, DALLE~\cite{dalle}, and others~\cite{dalle3,ediff,ernie-vilg2,sdxl,dai2023emu,pixart,hunyuan,cascade,playgroundv3,dreamomni,flux}. 


Recently, text-to-video (T2V) models~\cite{t2v1,t2v3,t2v4,t2v5,t2v6} have also advanced significantly. Specifically, Make-a-Video~\cite{t2v8} directly employs a cascade model that integrates spatial and temporal layers in the pixel space. Afterward, most works~\cite{t2v2,t2v7} perform the diffusion process in VAE-compressed latent space for efficiency. 
After that,  Sora~\cite{sora} generates highly realistic videos by scaling up training data and model size within the DIT architecture. Subsequently, many commercial T2V models~\cite{kling,moviegen,veo2} have also advanced significantly. Recently, several excellent DIT-based T2V works~\cite{cogvideox,mochi,ltx,hunyuanvideo,wan2025} have been open-sourced, promoting the research and application of T2V. Given the shared core capabilities between image and video editing, along with the rise of powerful T2V open-source models, building a unified instruction-based image and video editing model has become feasible.


\noindent\textbf{Image and Video Editing.} 
Editing methods can be classified into two categories: training-free~\cite{pnp,p2p,videop2p} and training-based~\cite{instructp2p,instructv2v}. Training-free editing has received widespread attention due to its lower dependence on training data and computational resources. Specifically, several works~\cite{edict, AIDI, ledits++, DDPM-inversion} have improved the inversion process~\cite{DDIM} to enhance editing consistency and efficiency. Other works~\cite{p2p, p2pzero, masactrl, pnp, ledits++, diffedit, llmga} explore modifying specific aspects of the diffusion model. For instance, Prompt-to-Prompt~\cite{p2p} adjusts cross-attention maps based on caption changes for editing. Moreover, video editing~\cite{lovecon,videop2p,fatezero,ditctrl,stablev2v,flatten,tokenflow,rfsolver} also uses similar methods.
However, training-free methods are difficult to optimize, have low success rates, and require extensive parameter tuning, making them less user-friendly. Moreover, training-free methods often rely on inversion, adding extra computation and making inference inefficient. In contrast, training-based methods are more suitable for commercial applications.


\begin{figure*}[t]
	\centering
 \resizebox{1\linewidth}{!}{
	\includegraphics[height=8cm]{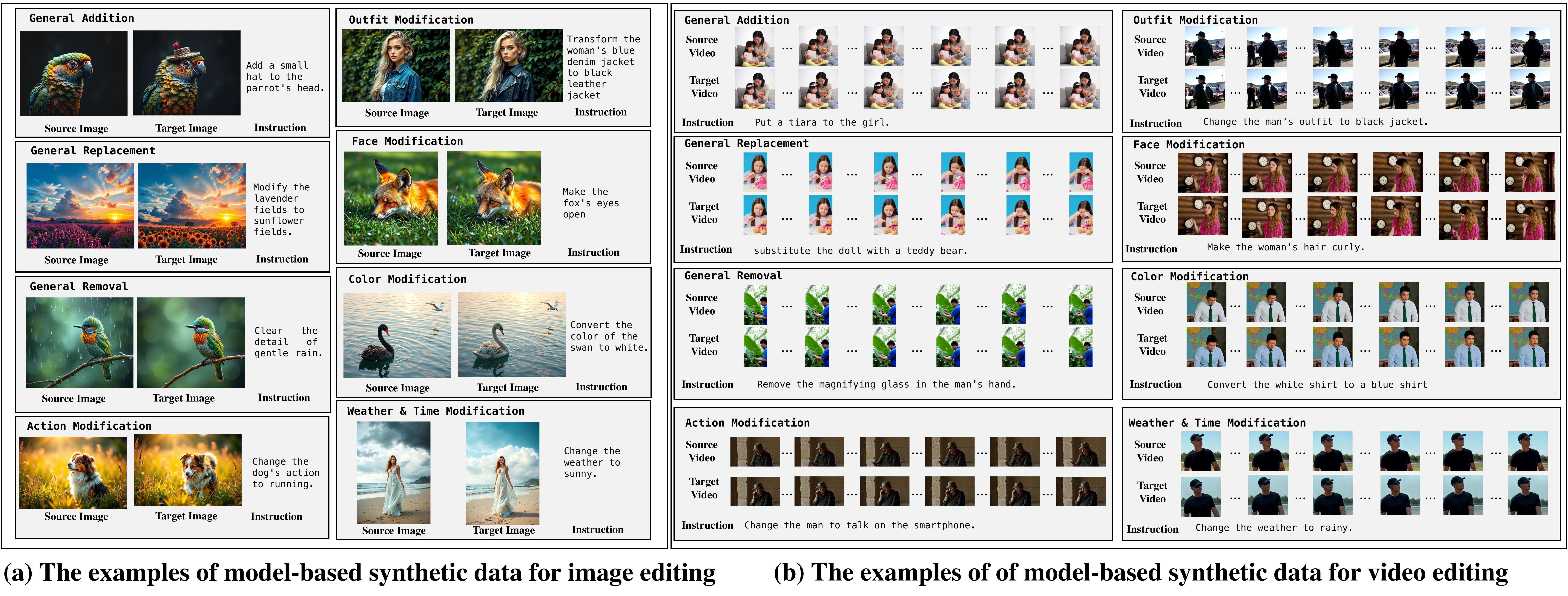}
 }
\vspace{-5mm}
	\caption{ 
The generative model-based synthetic data. (a) and (b) show examples of instruction-based image and video editing data. Their operations include general addition, removal, and replacement, as well as action, outfit, face, color, and weather \& time modifications.
}
\label{fig:ModelSyn}
\vspace{-7mm}
\end{figure*}

Among training-based editing tasks, instruction-based image editing~\cite{instructp2p,instructv2v,dreamomni,hq-edit,ace} has become popular for its flexible and simple interactive format. InstructP2P~\cite{instructp2p} is a pioneering work that introduced an instruction-based image editing dataset by fine-tuning GPT-3 and Prompt-to-Prompt~\cite{p2p} with SD 1.5~\cite{LDM}. Magicbrush~\cite{magicbrush} created a pipeline for generating editing image pairs using real images, where people used DALLE-2~\cite{dalle2} for inpainting. However, this manual process limits data scalability. Recently, various datasets~\cite{omniedit,seed-edit,ultraedit,anyedit} have been proposed, but editing quality and consistency remain limited by their base models. Video instruction-based editing is even less explored.
InsV2V~\cite{instructv2v}, like InstructP2P, fine-tunes a video model using T2I models and generates paired data via Prompt-to-Prompt~\cite{p2p}. However, it faces several challenges: low-quality, short (16-frame) video pairs from a weak base model, and insufficient filtering due to CLIP-based methods. To address these, \textbf{(1)} we propose a synthetic collage data pipeline for generating large-scale, high-quality pre-training data, covering most editing scenarios. \textbf{(2)} We introduce a generative model-based synthetic data pipeline to enhance specific attribute editing. Additionally, we present a feature-mixing scheme to adapt SOTA DIT-based models for generating high-quality editing data pairs in the pipeline, and apply strict MS-SSIM and GPT-4o evaluations for consistency, accuracy, and artifact-free data.

\vspace{-4mm}
\section{Methodology}
\vspace{-1mm}




In this paper, we propose DreamVE, a unified instruction-based image and video editing model, addressing two key challenges. \textbf{(1)} A major challenge in instruction-based editing is the lack of accurate data, particularly for video editing. In Sec.~\ref{sec:collage}, we discuss how collage-based data can be used to create extensive and accurate data covering primary editing tasks. 
However, collage-based data covers fewer corner cases in attribute editing. To this end, in Sec.~\ref{sec:generative}, we introduce the use of SOTA DIT-based generative models to generate corner editing cases for supplementation. Collage-based data is scalable, consistent, and accurate, making it ideal for pretraining and improving generalization, while generative model-based synthetic data helps fill gaps in corner cases. \textbf{(2)} Most existing research focuses on small, UNet-based SD models. In contrast, SOTA generative models use the DIT framework, flow-matching sampling, and distillation. They require new adaptations for data generation and training, rather than directly applying previous methods. We will discuss framework design in Sec.~\ref{sec:framework}. Moreover, We adopt a two-stage training scheme (Sec.~\ref{sec:framework}) to speed up DreamVE training and minimize dependence on video editing data.


\vspace{-1mm}
\subsection{Collage-based Synthetic Data}
\vspace{-1mm}
\label{sec:collage}

Instruction-based editing, particularly in videos, faces challenges due to the lack of accurate and enough training data. To address this, we propose two complementary synthetic data pipelines: one for generating collage-based data and another for creating data with generative models.

The examples of collage-based data for instruction-based image and video editing are shown in Fig.~\ref{fig:collageSyn}(a) and (b), respectively. Specifically, for the pipeline of collage-based image data, we use the saliency segmentation service to segment our collected large image dataset. Then, we collect foreground objects with high-confidence segmentation score to build our material database. We then use Qwen2-VL~\cite{qwen2-vl} to caption these foreground objects. Notably, our captions vary in detail: brief captions only mention the object subject, such as ``A fox.'', while detailed captions provide more information, like ``A fox with orange fur and white underbelly, looking to the side.''.

Then, we use foreground objects from the material database and background images to create the collage-based training data:
\textbf{(1)} For object removal, we randomly select and place several objects from the foreground database onto a background to create a source image. One object is then removed to form the target image, and the corresponding instruction is generated based on the removed object's caption. This approach helps train the model to accurately identify and remove the instructed object, even among multiple objects in the image.
\textbf{(2)} For object addition, we select a background image and randomly place a foreground object onto it to create the target image.
\textbf{(3)} For object replacement, we place multiple objects on a background to create the source image, then randomly remove one and replace it with another from the foreground database, resizing it to the same size and placing it in the same position.
\textbf{(4)} For quantity modification, we randomly place multiple instances of the same object on the background to create the source image, then randomly increase or decrease the number of objects in the target image.
\textbf{(5)} For color modification, we place colored geometric shapes on a background to create the source image, then randomly change the color of one shape to generate the target image and instruction.
\textbf{(6)} For size modification, we place multiple objects on the background to create the source image, then randomly scale one object by $\pm20\%$ to generate the target image.
\textbf{(7)} For segmentation and detection, we randomly select an object in the source image and apply a mask or bounding boxes in user-defined colors based on its alpha value.
\textbf{(8)} For background replacement, we create a source image by placing objects onto a new background, while the target image is the original image containing the objects. During source image generation, we apply brightness perturbations to help the model learn to adjust brightness according to different environments.
\textbf{(9)} For text editing (removal, addition, and replacement), the method of synthetic data is similar to object operations, except that the objects are replaced with randomly generated text.


Moreover, the video collage-based synthetic data pipeline is similar to the image-based one. Specifically, we mainly focus on basic operations like object removal, addition, and replacement (Fig.~\ref{fig:collageSyn}~(b)). First, we use saliency segmentation on the first frame of a filtered Panda-70M dataset~\cite{chen2024panda} to extract the foreground object. We collect high-confidence video clips and use SAM2~\cite{sam2} with the first-frame mask to generate foreground masks for the entire clip. Finally, we apply the same process as the image collage-based synthetic data pipeline to create source and target videos.

 
\vspace{-2mm}
\subsection{Generative Model-based Synthetic Data}
\vspace{-1mm}
\label{sec:generative}

Collage-based data offers extensive, diverse, consistent, and realistic editing data, but lacks sufficient attribute editing data, which can impact performance. To address this, we further propose a generative model-based synthetic pipeline.
While this pipeline has high computational demands, scalability issues, and lower consistency and diversity than collage-based data, it offers greater flexibility for attribute editing (Fig.~\ref{fig:ModelSyn}). Combining both approaches enables more comprehensive and robust editing.



Previously, data synthesis relied on SD1.5~\cite{LDM} or SDXL~\cite{sdxl} for generating paired image data. Recently, DIT-based models like FLUX~\cite{flux}, Hunyuan-Video~\cite{hunyuanvideo}, and Wan2.1~\cite{wan2025} offer improved detail generation and instruction following. In this paper, we use FLUX for image data, and for video editing, we chose Hunyuan-Video instead of Wan2.1-14B because although Wan2.1-14B provides slightly better quality, Hunyuan-Video is twice as fast due to CFG distillation. FLUX and Hunyuan-Video are based on flow-matching and the MM-DIT~\cite{sd3} architecture. Both models are distilled and incompatible with the previous synthetic data pipeline. To address this, we propose a feature-mixing approach for data generation. Specifically, we first use Qwen2-VL~\cite{qwen2-vl} to generate concise source captions for image and video datasets. Next, GPT-4o generates diverse editing instructions and target captions. The editing types for both image and video include general addition, replacement, and removal, as well as action, outfit, face, color, and weather \& time modification.



We treat the source and target captions as a pair and input them into Shuttle-3-diffusion~\cite{shuttle-3-diffusion} (a variant model of FLUX) or Hunyuan-Video~\cite{hunyuanvideo} for parallel generation with a batch size of $2$ (\ie, source and target generation branches) to generate the source and target image/video editing data. Notably, our method can also be applied to Wan2.1 (see supplementary material), demonstrating its strong generalization. In the first $N$ layers, we mix the source branch’s noise features from the attention module into the target branch's attention computation at corresponding positions as follows:
\vspace{-2mm}
\begin{equation}
\operatorname{Attn}_{tar}(\Q,\K,\V)=\operatorname{softmax}\left(\frac{\Q \K^{\top}}{\sqrt{d}}\right) \V,
\end{equation}
where $Q=[Q^n_{tar}; Q^t_{tar}]$, $K=[K^n_{tar}; K^t_{tar}; K^n_{src}]$, and $V=[V^n_{tar}; V^t_{tar}; V^n_{src}]$. $Q^t_{tar}$, $K^t_{tar}$, and $V^t_{tar}$ are the text features from the target branch, while $Q^n_{tar}$, $K^n_{tar}$, and $V^n_{tar}$ are the noise features from the target branch. $K^n_{src}$ and $V^n_{src}$ are the noise features from the source branch at the same layer as the $K^n_{tar}$ and $V^n_{tar}$. $[;]$ indicates token (or called length) dimension concatenation.


Notably, the source branch does not perform feature mixing, and the target branch only mixes with the noise features from the source branch during the attention computation in the first $N$ layers. Additionally, both the source and target branches use the same initial Gaussian noise to ensure consistency in the details of the generated paired data.


\begin{figure}[t]
	\centering
 \resizebox{1\linewidth}{!}{
	\includegraphics[height=8cm]{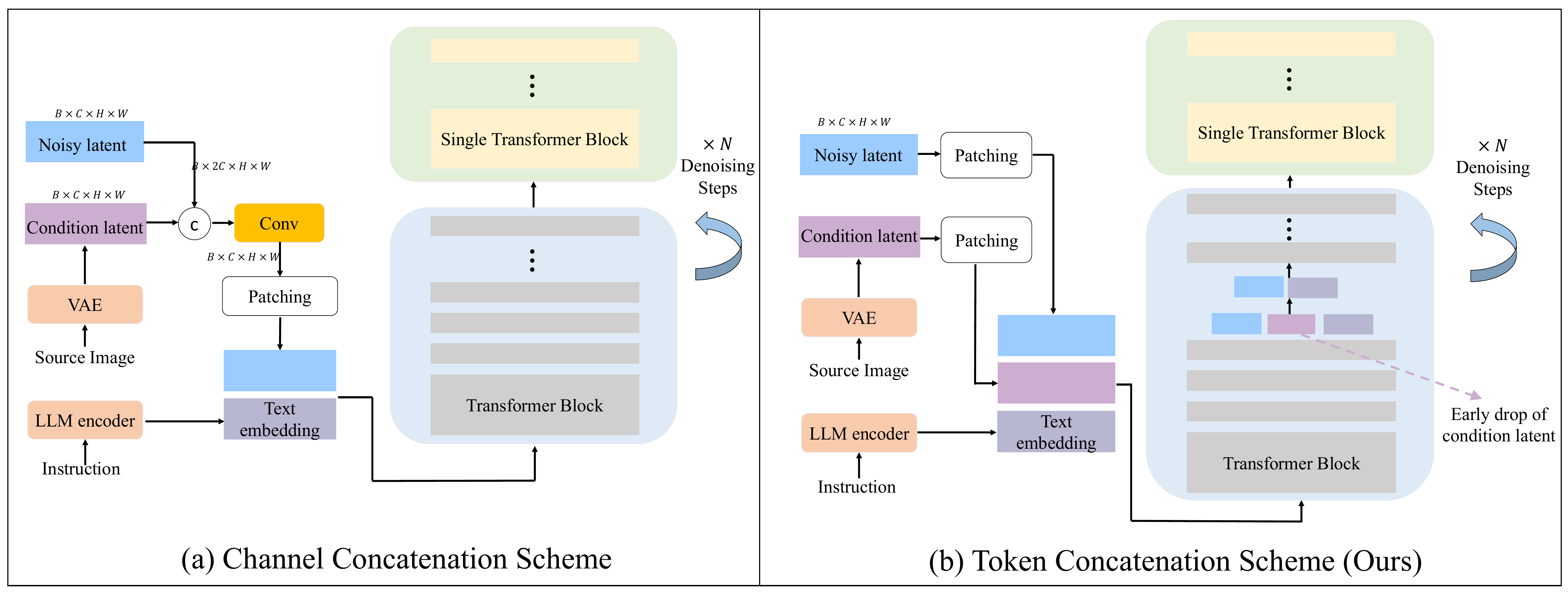}
 }
 \vspace{-4mm}
	\caption{ 
The comparison of two source image injection schemes. \textbf{(a)} The noisy latent and condition latent are concatenated along the channel dimension, followed by a convolutional layer to reduce the channels to match the noisy latent. \textbf{(b)} Our adopted token concatenation scheme. For efficiency, the condition latent is used only in the first four layers and is then dropped out for the remaining layers.
}
\label{fig:framework}
\vspace{-4mm}
\end{figure}

After generating synthetic data, we filter out samples with poor instruction adherence, low consistency, or visible artifacts. For image data, we calculate the MS-SSIM between the source and target images, discarding low-consistency pairs, then use GPT-4o to check instruction adherence and artifacts. For video data, we uniformly sample 5 frames from both the source and target video clips, compute the average MS-SSIM, and discard those below the threshold. The remaining video clips are reviewed by GPT-4o, and if four or more pass, the video clip is considered valid.

\begin{figure*}[t]
	\centering
 \resizebox{1\linewidth}{!}{
	\includegraphics[height=4cm]{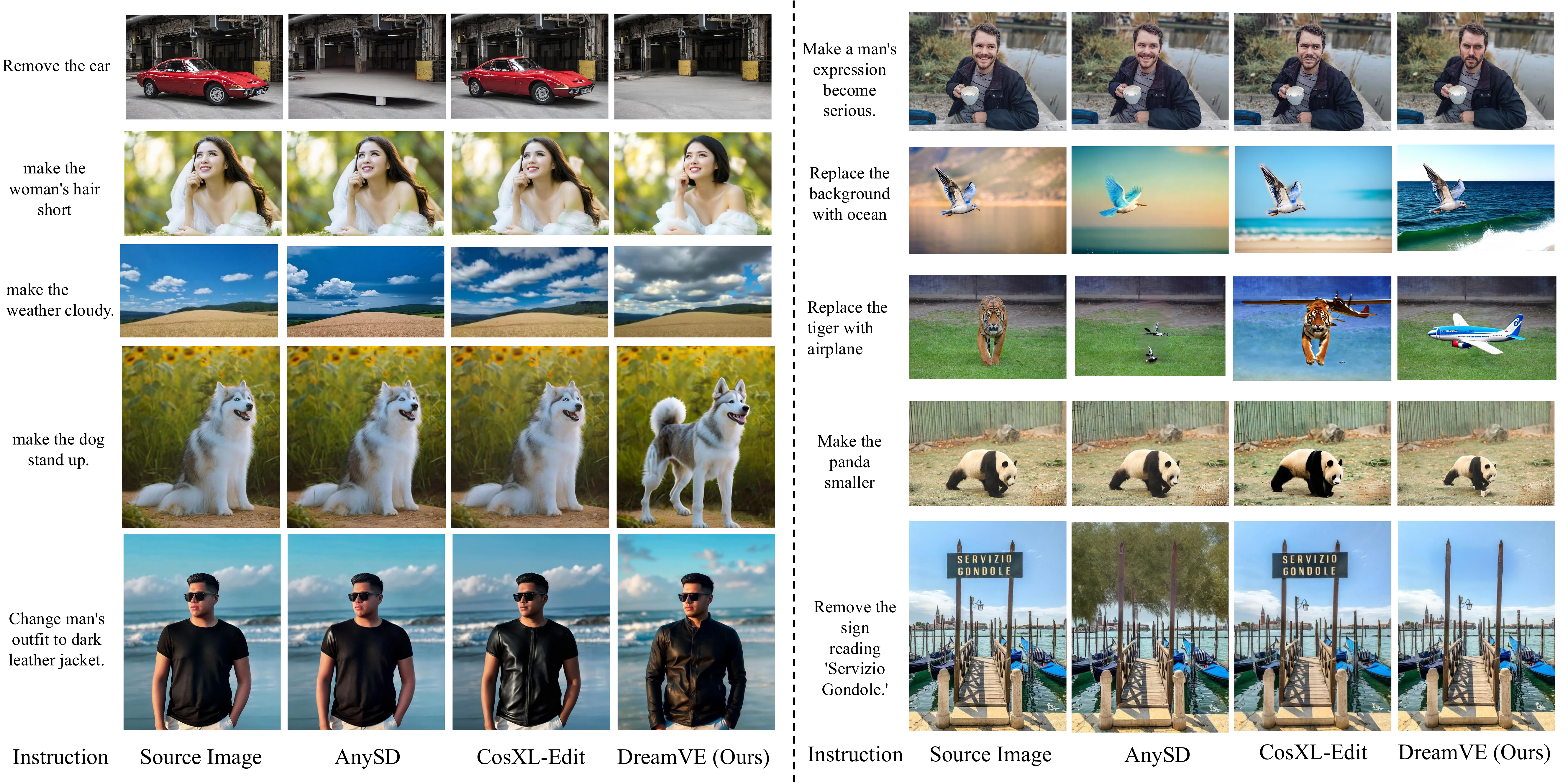}
 }
  \vspace{-5mm}
	\caption{Comparison on \textbf{instruction-based image editing}. DreamVE achieves more precise editing results over diverse editing tasks compared to the latest models (CosXL-Edit~\cite{cosxl} and AnySD~\cite{anyedit}).  }
	\label{fig:ip2p}
 \vspace{-5mm}
\end{figure*}

Furthermore, we prioritize data diversity during generation. For image data, we use a $512 \times 512$ resolution and divide images into $31$ bins with aspect ratios uniform ranging from $4:1$ to $1:4$. Feature mixing layers in generative-model based image editing generation is set to $N=18$. For video data, frames are randomly sampled from $[73, 77, 81, 85]$, and resolutions are chosen randomly from $[320 \times 544, 384 \times 480, 416 \times 416, 480 \times 384, 544 \times 320]$. The $N$ is set to $13$. After that, we apply the super-resolution model, DiffIR~\cite{diffir}, for $2\times$ upsampling on video data.


\vspace{-1mm}
\subsection{Framework Design and Training}
\label{sec:framework}

We trained our unified instruction-based editing model on the Hunyuan-Video. As shown in Fig.~\ref{fig:framework}, we compared two methods for injecting source image guidance: \textbf{(1)} In Fig.~\ref{fig:framework}~(a), we concatenate the source condition latent and noisy latent along the channel dimension, then apply a convolutional layer to compress the result back to the original number of channels in the noisy latent. \textbf{(2)} In Fig.~\ref{fig:framework}~(b), based on the MM-DIT architecture, we concatenate the latents along the token dimension. The source condition latent is used only in the first four transformer blocks, after which it is discarded.  While both schemes have similar runtime, the second scheme offers significantly better editing performance and supports future multi-task extensions~\cite{dreamomni,omnigen,magicmirror,liu2024generative}. Therefore, the second scheme is adopted as the editing framework of our DreamVE. For our training strategy, we first train on image editing, followed by video editing. The prior knowledge from image editing accelerates and improves video editing training. This reduces the need for large-scale video data and lowers computational costs.


\begin{figure*}[t]
    \centering
    \resizebox{1\linewidth}{!}{
        \includegraphics[height=3.6cm]{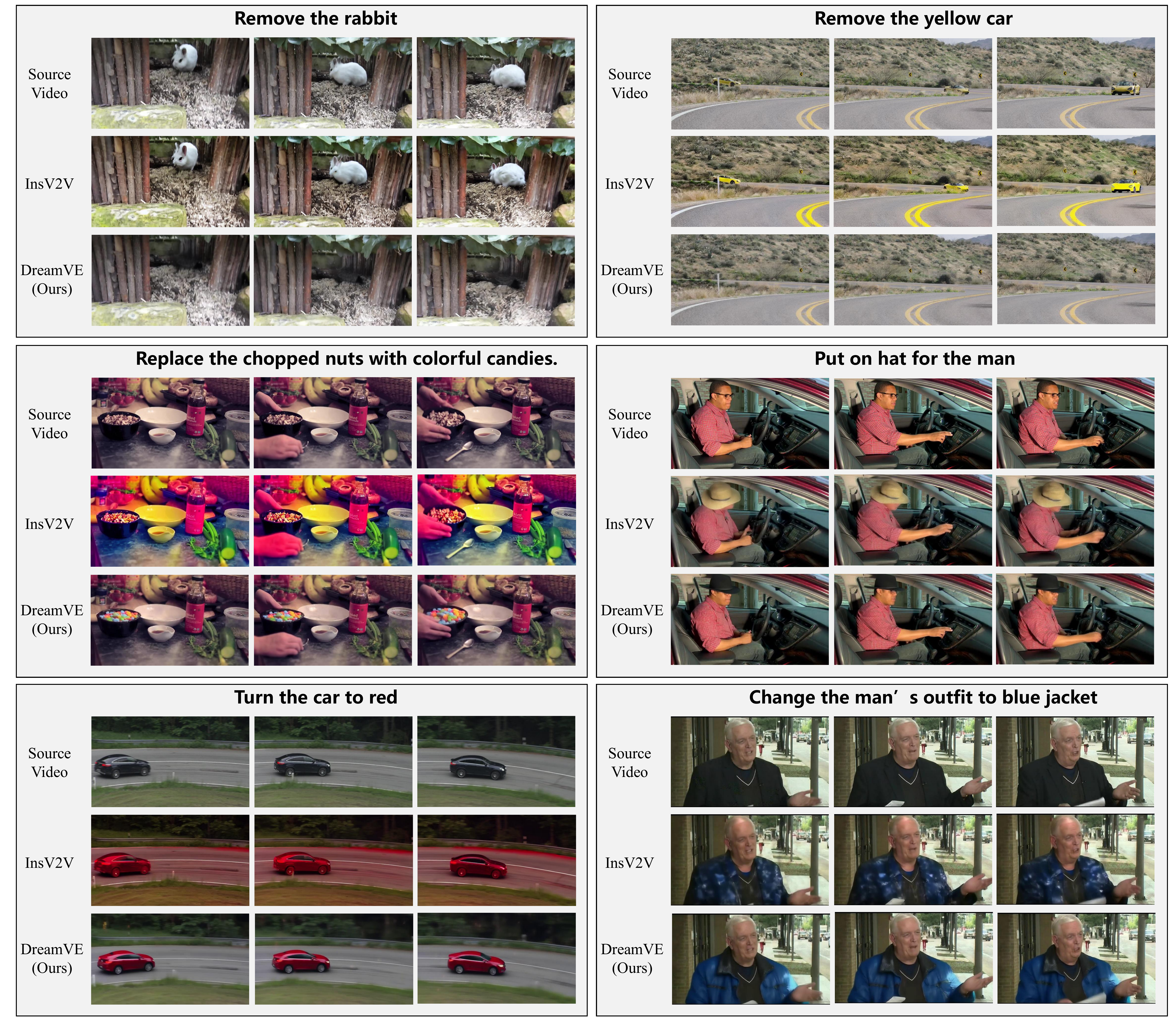}
    }
    \vspace{-5mm}
    \caption{Visual comparison on \textbf{instruction-based video editing}: our DreamVE significantly outperforms the related work on diverse editing tasks in terms of instruction.  }
    \label{fig:iv2v}
    \vspace{-4mm}
\end{figure*}

We trained DreamVE using Lora on Hunyuan-Video in four stages:
\textbf{(1)} We trained on $5M$ collage-based image editing data with a batch size of $1024$ and a learning rate of $1 \times 10^{-4}$.
\textbf{(2)} We then trained on $199,064$ generative image synthetic data and $1M$ collage synthetic data, using a batch size of $1024$ and the same learning rate.
\textbf{(3)} We trained on $200K$ collage video editing data, $199,064$ generative image synthetic data, and $200K$ collage image editing data, with a batch size of $64$ and a learning rate of $4 \times 10^{-5}$.
\textbf{(4)} We trained on $28,008$ generative video synthetic data, $30K$ collage video editing data, $199,064$ generative image synthetic data, and $200K$ collage image editing data, with a batch size of $64$ and the same learning rate. We train our models on Nvidia H20.


\vspace{-3mm}
\section{Experiments}
\vspace{-2mm}

\noindent\textbf{Evaluation on Instruction-based Image Editing.} 
To evaluate the model's performance, we collaborated with frontline engineers specializing in editing to create the DreamVE image test set, reflecting real-world applications. We compared our approach with classic instruction-based image editing methods, such as InstructP2P~\cite{instructp2p}, as well as competitive open-source solutions from leading companies, including MGIE~\cite{MGIE} and CosXL-Edit~\cite{cosxl}. As shown in Tab.~\ref{tab:metric-image-edit}, our results significantly outperform the others in CLIP-T, CLIP-I, and MS-SSIM, demonstrating better adherence to editing instructions while maintaining consistency in non-edited areas. DreamVE also outperforms AnySD~\cite{anyedit}, demonstrating the better quality of our data synthesis pipeline.

\begin{wraptable}{r}{0.48\linewidth}
	\small
	\centering
	\vspace{-4mm}
  \caption{Results on instruction-based image editing.}
  \vspace{-1mm}
  \resizebox{1\linewidth}{!}{
    \begin{tabular}{l|ccc}
    \toprule[0.2em]
    \textbf{Method} & \textbf{CLIP-T}$\uparrow$ & \textbf{CLIP-I}$\uparrow$ & \textbf{MS-SSIM}$\uparrow$ \\
    \midrule
    InstructP2P~\cite{instructp2p} & 0.2945  & 0.8598  & 0.7457  \\
    Magicbrush~\cite{magicbrush} & 0.3056  & 0.8972  & 0.7850  \\
    HQ-edit~\cite{hq-edit} & 0.2777  & 0.7738  & 0.3349  \\
    MGIE~\cite{MGIE}  & 0.3004  & 0.8802  & 0.7413  \\
    CosXL-Edit~\cite{cosxl} & 0.3052  & 0.8996  & 0.8021  \\
    UltraEdit~\cite{ultraedit} & 0.3023  & 0.8875  & 0.7699  \\
    AnySD~\cite{anyedit} & 0.3038  &  0.8902 &  0.7868 \\
    \midrule
    DreamVE (Ours) & \textbf{0.3143}  & \textbf{0.9089}  & \textbf{0.8246}  \\
    \bottomrule[0.2em]
    \end{tabular}%
    }
    \vspace{-3mm}
  \label{tab:metric-image-edit}%
\end{wraptable}

Additionally, visual comparisons are shown in Fig.~\ref{fig:ip2p}. Our method was evaluated on various editing tasks, demonstrating strong generalization, better instruction adherence, and high-quality edits. These results highlight the importance of training data for the performance of the editing model.  Our two synthetic data pipelines cover diverse editing scenarios, enabling scalable high-quality training data to improve the model's performance and generalization. They also show the effectiveness of our designed editing framework based on the latest MM-DIT base model.



\begin{wraptable}{r}{0.52\linewidth}
    \small
    \centering
    \vspace{-1mm}
    \caption{Results on instruction-based video editing. }
    \vspace{-2mm}
    \resizebox{1\linewidth}{!}{
        \begin{tabular}{l|ccc}
            \toprule[0.2em]
            \textbf{Method} & \textbf{CLIP-T}$\uparrow$ & \textbf{CLIP-I}$\uparrow$ & \textbf{MS-SSIM}$\uparrow$ \\
            \midrule
            RF-Solver-Edit~\cite{rfsolver} & 0.2735  & 0.8859  & 0.8782  \\
            TokenFlow~\cite{tokenflow} & 0.2720  & 0.8807 & 0.8745  \\
            InsV2V~\cite{instructv2v} & 0.2749  & 0.8836  & 0.8790  \\
            \midrule
            DreamVE (Ours) & \textbf{0.2846}  & \textbf{0.8970}  & \textbf{0.8913}  \\
            \bottomrule[0.2em]
        \end{tabular}%
    }
    \label{tab:metric-video-edit}%
    \vspace{-4mm}
\end{wraptable}


\noindent\textbf{Evaluation on Instruction-based Video Editing.} 
Currently, instruction-based video editing lacks a comprehensive benchmark for evaluating the performance of the editing model. To this end, we developed the DreamVE video test set, which covers a wide range of video editing scenarios. Besides, using CLIP-I, CLIP-T, and MS-SSIM, we assess instruction adherence and consistency by uniformly sampling five frames from each video clip and averaging the results. We compare InsV2V~\cite{instructv2v}, TokenFlow~\cite{tokenflow}, and the training-free Hunyuan-Video-based method RF-Solver-Edit~\cite{rfsolver}.

\begin{wraptable}{r}{0.28\linewidth}
    \small
    \centering
    \vspace{-5mm}
    \caption{Video editing runtime (same base model) }
    \vspace{-2mm}
    \resizebox{1\linewidth}{!}{
        \begin{tabular}{l|c}
            \toprule[0.2em]
            \textbf{Method} & \textbf{Runtime (ms)}  \\
            \midrule
            RF-Solver-Edit~\cite{rfsolver} & 61.2    \\
            \midrule
            DreamVE (Ours) & 28.2    \\
            \bottomrule[0.2em]
        \end{tabular}%
    }
    \label{tab:runtime}%
    \vspace{-4mm}
\end{wraptable}

In Tab.~\ref{tab:metric-video-edit} and Fig.~\ref{fig:iv2v}, DreamVE  outperforms them. Notably, DreamVE outperforms the training-free RF-Solver-Edit and is $2.2\times$ faster, despite using the same base model.  Our improvement comes from: \textbf{(1)} A scalable collage-based pipeline enabling diverse, generalizable data. \textbf{(2)} A generative approach with feature mixing for fine-grained editing, powered by T2V models. \textbf{(3)} An efficient instruction-based editing framework built on DIT. More visual results are in the supplementary.

\begin{wraptable}{r}{0.58\linewidth}
    \small
    \centering
    \vspace{-5mm}
    \caption{The comparison of editing frameworks. }
    \vspace{-2mm}
    \resizebox{0.9\linewidth}{!}{
        \begin{tabular}{l|c|ccc}
            \toprule[0.2em]
            \textbf{Method} & \textbf{Runtime (ms)} & \textbf{CLIP-T}$\uparrow$ & \textbf{CLIP-I}$\uparrow$ & \textbf{MS-SSIM}$\uparrow$ \\
            \midrule
            Channel Concat & 27.4  & 0.3116 & 0.9067 & 0.8203 \\
            Token Concat & 28.2  & 0.3149 & 0.9098 & 0.8250 \\
            \bottomrule[0.2em]
        \end{tabular}%
    }
    \vspace{-3mm}
    \label{tab:quan_expframework}%
\end{wraptable}

\begin{figure*}[t]
\begin{minipage}[t]{0.47\linewidth}
\centering
\includegraphics[width=0.78\textwidth]{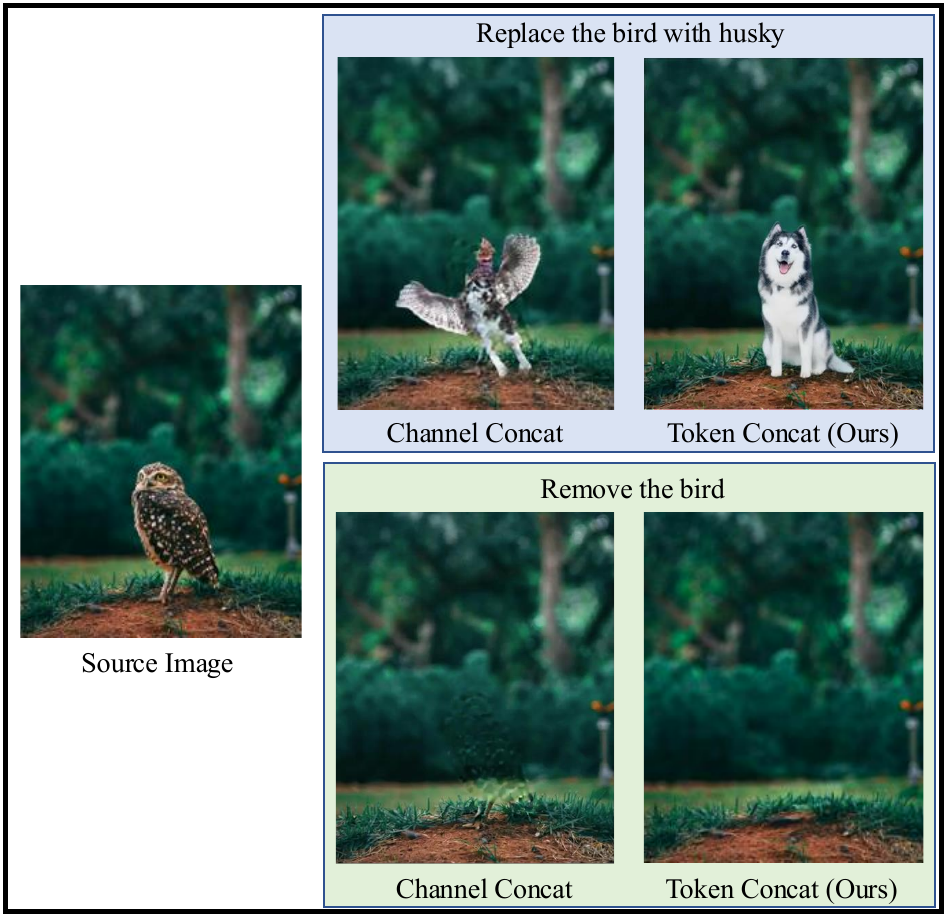}
\vspace{-1mm}
\caption{The comparison of frameworks.}
\label{fig:expframework}
 \end{minipage}%
\hspace{3mm}
\begin{minipage}[t]{0.48\linewidth}
\centering
\includegraphics[width=0.78\textwidth]{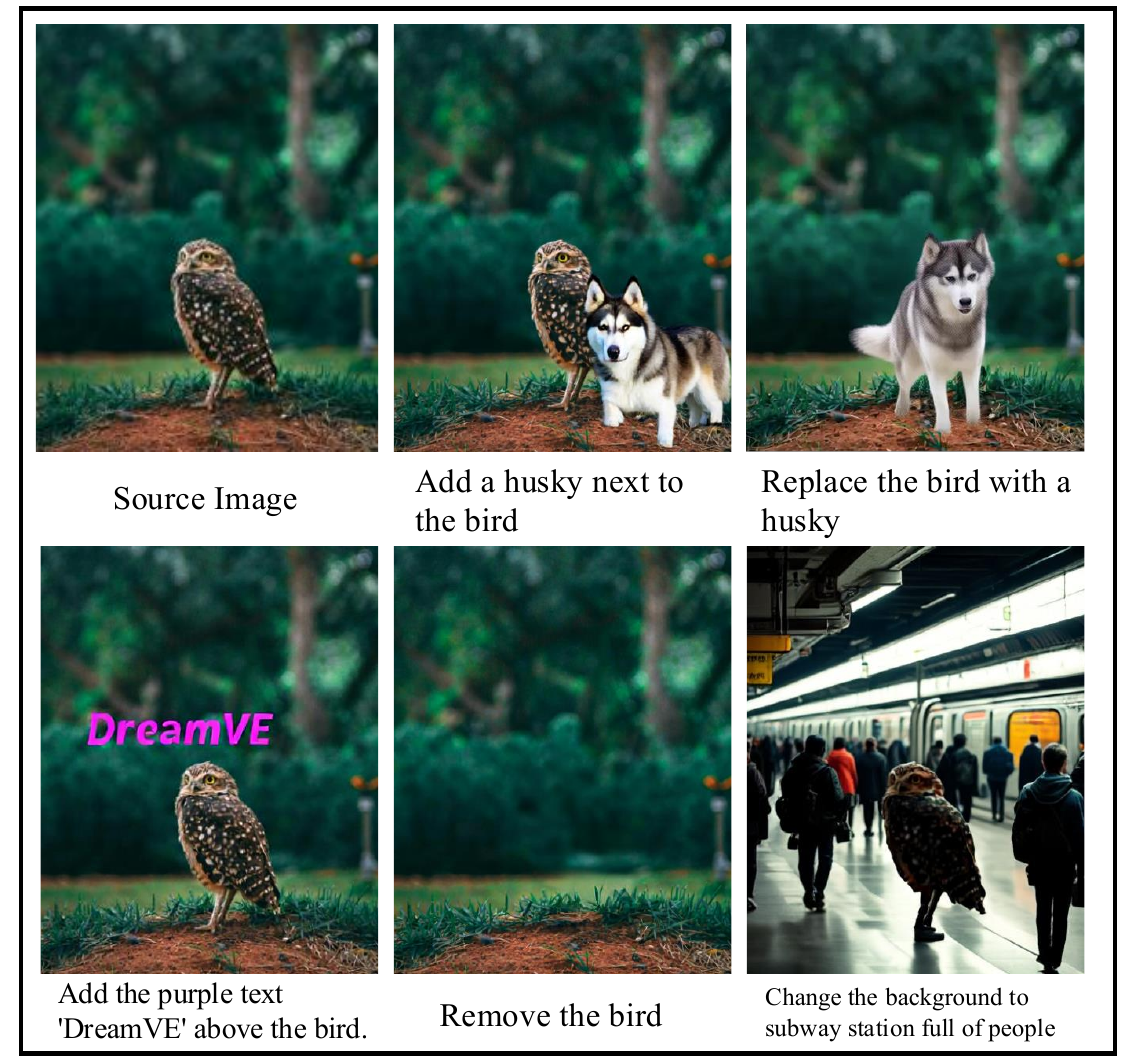}
\vspace{-1mm}
\caption{DreamVE only trained on collage data.}
\label{fig:exp-collage}
\end{minipage}
\vspace{-5mm}   
\end{figure*}

\noindent\textbf{Framework Design.} 
Our DreamVE is built on the powerful T2V model Hunyuan-Video, which utilizes the MM-DIT architecture, Flow Matching sampling, and CFG distillation. As shown in Fig.~\ref{fig:framework}, we designed two editing frameworks to inject source image/video information into the network: one is the classic approach of channel concatenation between source condition latent and noisy latent, while the other leverages MM-DIT's property by token concatenation between noisy latent and source condition latent. The source condition latent undergoes early dropout after passing the first four layers. For simplicity, both frameworks were trained only for image editing (\ie, phase one and two in Sec.~\ref{sec:framework}).
As shown in Tab.~\ref{tab:quan_expframework}, we performed quantitative testing on the DreamVE image test set, and it was evident that our token concatenation with early drop scheme significantly outperforms channel concatenation. Additionally, we tested the single denoising time for both schemes on the H20 GPU, and the results showed that the time required for both schemes is almost identical. Moreover, we presented the editing results of both schemes in Fig.~\ref{fig:expframework}, where it is clear that the channel concatenation approach suffers from severe artifacts. Therefore, we adopted the token concatenation with early dropout scheme in DreamVE.

\begin{wraptable}{r}{0.66\linewidth}
    \small
    \centering
    \vspace{-5mm}
    \caption{The effect of synthetic training data.}
    \vspace{-2mm}
    \resizebox{1\linewidth}{!}{
        \begin{tabular}{l|cc|ccc}
            \toprule[0.2em]
            \textbf{Method} & \makecell{\textbf{Collaged-based} \\ \textbf{data}} & \makecell{\textbf{Generative model}\\  \textbf{based data}} & \textbf{CLIP-T}$\uparrow$ & \textbf{CLIP-I}$\uparrow$ & \textbf{MS-SSIM}$\uparrow$ \\
            \midrule
            DreamVE-V1  & \checkmark     &       & 0.3077 & 0.9034 & 0.8193 \\
            DreamVE-V2 &       & \checkmark     & 0.3064 & 0.9017 & 0.8065 \\
            DreamVE-V3 (Ours) & \checkmark     & \checkmark     & 0.3149 & 0.9098 & 0.8250 \\
            \bottomrule[0.2em]
        \end{tabular}%
    }
    \vspace{-4mm}
    \label{tab:quan_collage}%
\end{wraptable}

\noindent\textbf{The Effectiveness of Training Data.} 
We aim to explore how collage-based synthetic data and generative model-based synthetic data affect the editing performance of DreamVE, focusing on image editing training for experimental convenience. As shown in Tab.~\ref{tab:quan_collage}, we progressively add these two data types to examine their roles.
We conduct three experiments:
\textbf{(1)} DreamVE-V1: trained with only collage-based data,
\textbf{(2)} DreamVE-V2: trained with only generative model-based data,
\textbf{(3)} DreamVE-V3: trained with both data types.
While DreamVE-V2 offers a broader range of editing types than DreamVE-V1, its performance is slightly worse in quantitative metrics. This is due to the smaller amount of training data in DreamVE-V2, which limits the model’s generalization and responsiveness. Moreover, collage-based synthetic data also has better consistency than generative model-based synthetic data.
When we combine both data types in DreamVE-V3, we see a significant improvement. This suggests the two data types are complementary. The collage-based data provides a solid foundation, and generative model-based data fine-tunes corner editing cases, boosting overall performance.
Furthermore, in Fig.~\ref{fig:collageSyn}, we show the editing results of DreamVE-V1, trained only with collage-based data. 
Despite the inherent messiness of collage-based data, the model trained on it performs surprisingly well in responding to the primary editing tasks. This can be attributed to T2V model’s denoising process, which naturally guides the generated content toward the realistic domain.

\vspace{-3mm}
\section{Conclusion}
\vspace{-2mm}

We introduce DreamVE, a unified instruction-based model for image and video editing that tackles the challenges of limited training data and presents a novel framework and two-stage training scheme. Specifically, we propose a collage-based synthetic data pipeline, which is scalable, accurate, and diverse, helping DreamVE develop core editing and generalization abilities. Furthermore, to cover corner attribute editing cases not handled by the collage-based data, we supplement it with a generative model-based synthetic data pipeline using the SOTA T2I and T2V models, creating editing data with better quality and editing accuracy than prior methods. To adapt these SOTA T2I and T2V models for data generation, we introduce a feature-mixing scheme. For the framework, we adopt a token concatenation with early-drop scheme to balance editing responsiveness and consistency while keeping high efficiency. Extensive experiments validate the DreamVE's effectiveness.

{
\small

\bibliographystyle{plain}
\bibliography{main}
}


\appendix



The overview of the supplementary materials:

\textbf{(1)} We discuss current limitations of DreamOmni and provide future improvements (Sec.~\ref{sec:limation}).

\textbf{(2)} We discuss the border impacts of our DreamVE (Sec.~\ref{sec:impact}).

\textbf{(3)} We apply our feature mixing method to Wan2.1 and can generate high-quality data, demonstrating the strong generalization capability of our approach. (Sec.~\ref{sec:general}).

\textbf{(4)} We compared the acceleration and performance improvement brought by our two-stage training approach for video editing training (Sec.~\ref{sec:twostage}).

\textbf{(5)} supplementary (Sec.~\ref{sec:visual}).

\textbf{(6)} We provide more details about the collage-based synthetic data pipeline (Sec.~\ref{sec:data}).

\textbf{(7)} We provide more details about how to make instruction data (Sec.~\ref{sec:gpt4o}).

\textbf{(8)} We provide the link to coherent video demonstrations to facilitate a more intuitive comparison (Sec.~\ref{sec:video}).

\section{Limitations}
\label{sec:limation}

Although DreamVE currently achieves state-of-the-art performance in both video and image editing, and stands out as the most efficient among methods built on similar foundation models, there are still some limitations and areas for improvement:
\textbf{(1)} While Wan-14B offers superior generation quality compared to HunYuan-video, we chose to use HunYuan-video~\cite{hunyuanvideo} as the base model in this work due to practical constraints. Wan2.1~\cite{wan2025} lacks a CFG-distilled version, resulting in inference speeds that are approximately twice as slow. With future improvements in computational resources, adapting Wan-14B could further enhance performance.
\textbf{(2)} The model occasionally struggles with generating realistic hands — a known challenge across the community. Addressing this issue will require further research and targeted advancements.

\section{Broader impacts}
\label{sec:impact}
\textbf{Positive societal impacts}. \textbf{(1)} We propose an efficient two-stage training strategy, a high-performance video editing framework, and a cost-effective method for generating video-image data. Since image and collage data are significantly cheaper to produce than video editing data, our pipeline greatly reduces the reliance on expensive video editing data while achieving superior performance.
\textbf{(2)} In addition, our approach explores the possibility of a unified framework for both video and image generation/editing.
\textbf{(3)} We have also constructed a set of high-quality image and video editing datasets, which can serve as valuable resources to support future research.
 
\textbf{Negative societal impacts}. The capabilities of image and video editing could potentially be misused for illegal purposes, such as creating forged content for fraud or spreading misinformation.

\section{The Generalizability of Synthetic Data Pipeline }
\label{sec:general}

\begin{figure}[t]
	\centering
 \resizebox{1\linewidth}{!}{
	\includegraphics[height=8cm]{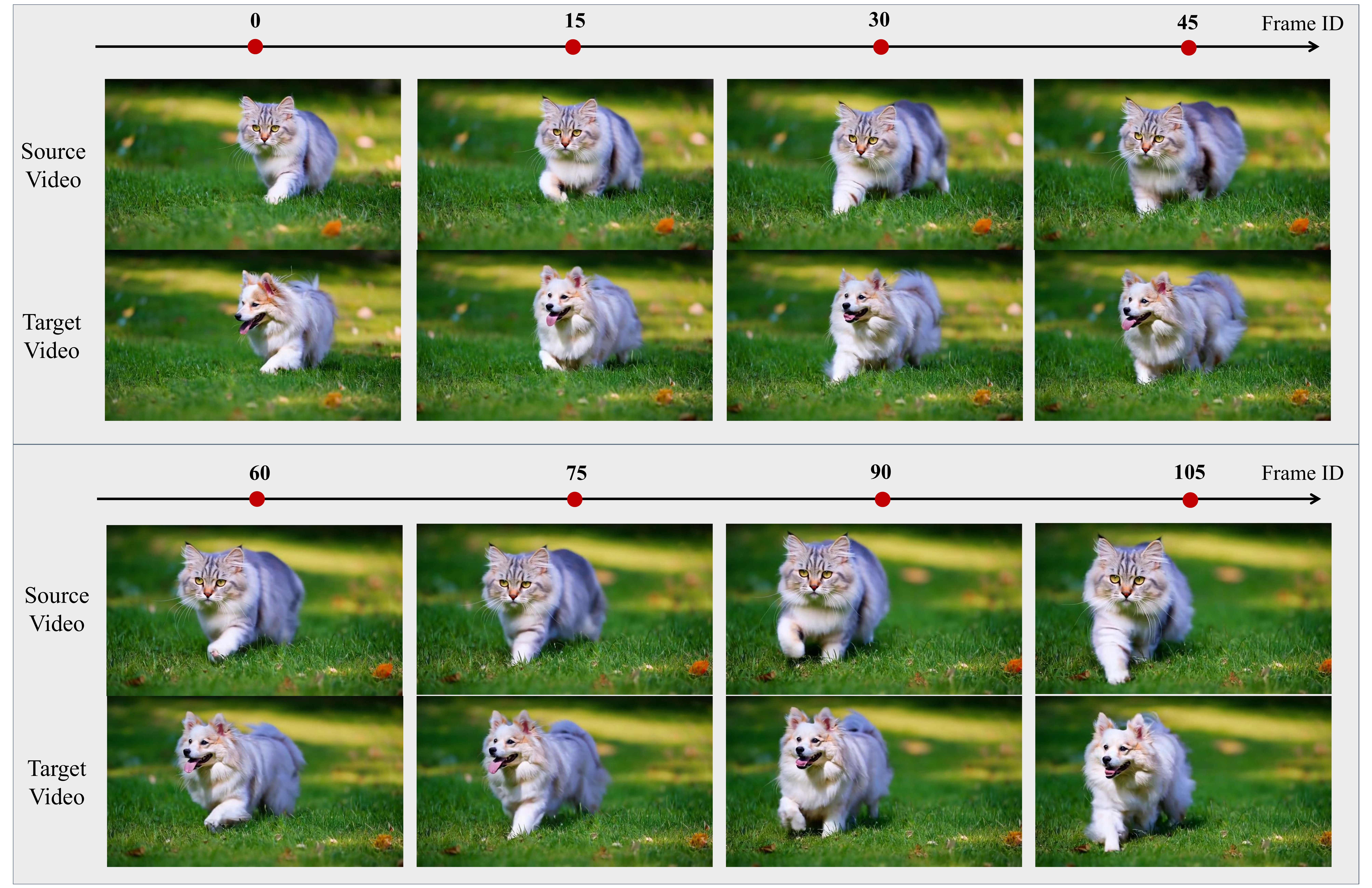}
 }
 \vspace{-4mm}
	\caption{ 
The visualization results of the data produced by applying our feature mixing scheme on Wan2.1.
}
\label{fig:wan}
\vspace{-4mm}
\end{figure}

Our generative model-based synthetic data method can also be applied to Wan2.1~\cite{wan2025}. The difference between Wan2.1 and Hunyuan-Video lies in that Wan2.1 uses cross-attention and self-attention, whereas Hunyuan-Video~\cite{hunyuanvideo} uses MM-DIT. For Wan2.1, in the first $2$ layers, we mix the source branch’s noise features from the self-attention module into the target branch's self-attention computation at corresponding positions as follows:

\vspace{-2mm}
\begin{equation}
\operatorname{Attn}_{tar}(\Q,\K,\V)=\operatorname{softmax}\left(\frac{\Q \K^{\top}}{\sqrt{d}}\right) \V,
\end{equation}
where $Q=Q^n_{tar}$, $K=[K^n_{tar}; K^n_{src}]$, and $V=[V^n_{tar}; V^n_{src}]$. $Q^t_{tar}$, $K^t_{tar}$, and $Q^n_{tar}$, $K^n_{tar}$, and $V^n_{tar}$ are the noise features from the target branch. $K^n_{src}$ and $V^n_{src}$ are the noise features from the source branch at the same layer as the $K^n_{tar}$ and $V^n_{tar}$. $[;]$ indicates token (or called length) dimension concatenation.

Notably, the source branch does not perform feature mixing, and the target branch only mixes with the noise features from the source branch during the attention computation in the first $2$ layers. Additionally, both the source and target branches use the same initial Gaussian noise to ensure consistency in the details of the generated paired data.

As shown in Fig.~\ref{fig:wan}, our method remains effective on Wan2.1, capable of producing accurate and high-quality paired editing data. This further demonstrates the strong generalization ability of our feature mixing scheme, which can be applied to different models.

\section{Validation on Two-Stage Training Scheme }
\label{sec:twostage}

\begin{figure}[t]
	\centering
 \resizebox{0.8\linewidth}{!}{
	\includegraphics[height=8cm]{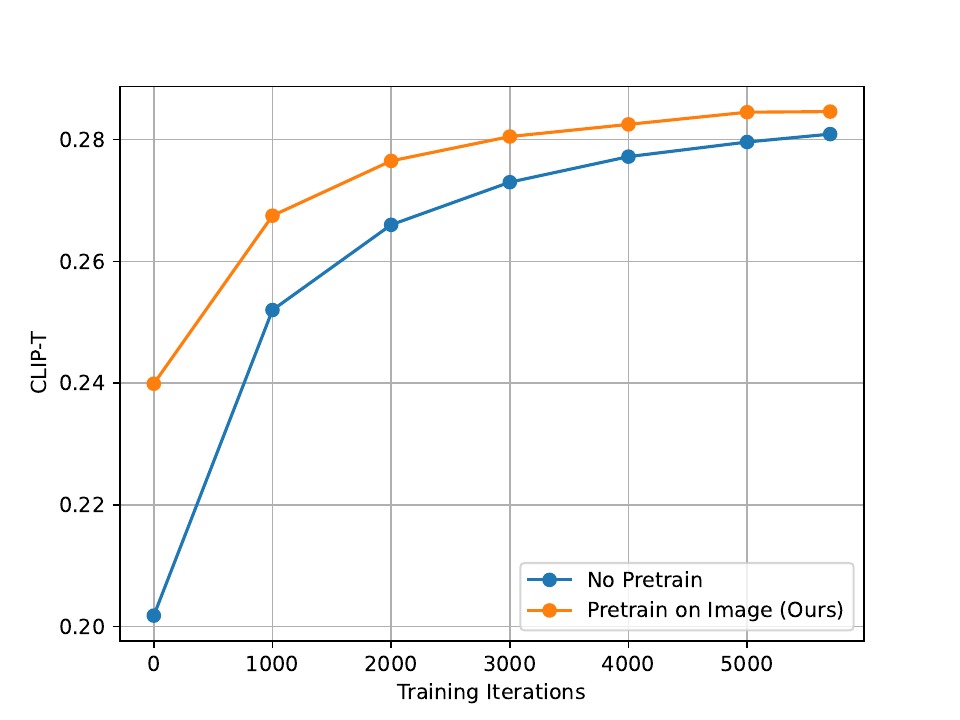}
 }
 \vspace{-4mm}
	\caption{ 
Validation of the two-stage training scheme.
}
\label{fig:two-stage}
\vspace{15mm}
\end{figure}

As shown in Fig.~\ref{fig:two-stage}, we compare two schemes: training a video editing model directly on the T2V base model, and pretraining it on image editing tasks before fine-tuning for video editing. The results demonstrate that models pretrained on image editing not only start from a stronger baseline, but also achieve comparable performance with nearly half the training iterations. This highlights that image pretraining can significantly accelerate training convergence and improve the performance of video editing models. Importantly, the cost of acquiring and training on image editing data is substantially lower than that of video editing, effectively doubling the training efficiency. This validates the effectiveness of our Two-Stage Training Scheme. Moreover, the model is capable of unifying both image and video editing within a single framework.

\section{More Visual Comparisons for Video Editing }
\label{sec:visual}
\begin{figure*}[t]
	\centering
 \resizebox{0.8\linewidth}{!}{
	\includegraphics[height=4cm]{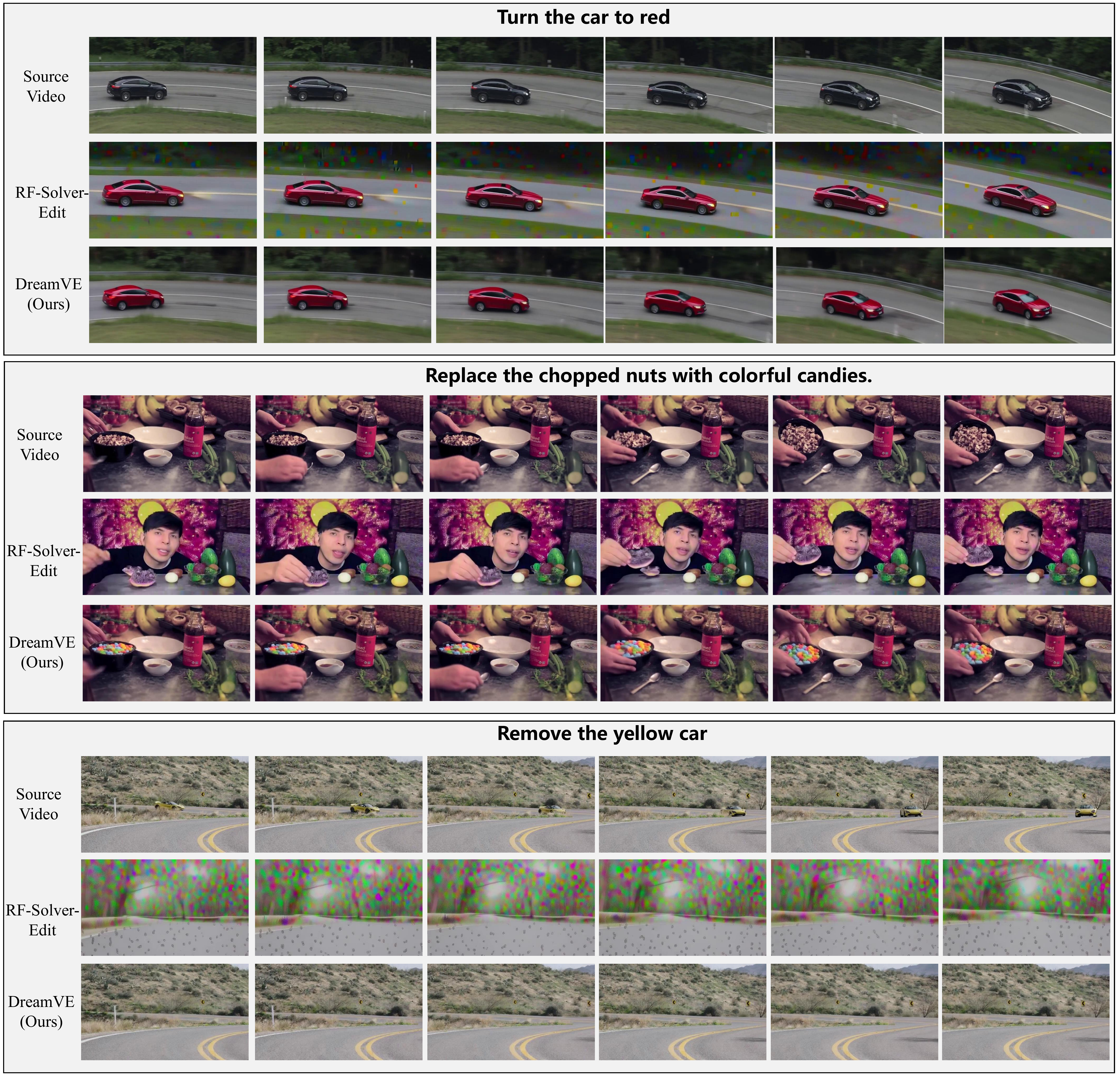}
 }
 \vspace{-2mm}
	\caption{The collage-based synthetic data pipeline (take object replacement as an example).
 } 
	\label{fig:vis1}
\end{figure*}

\begin{figure*}[t]
	\centering
 \resizebox{0.8\linewidth}{!}{
	\includegraphics[height=4cm]{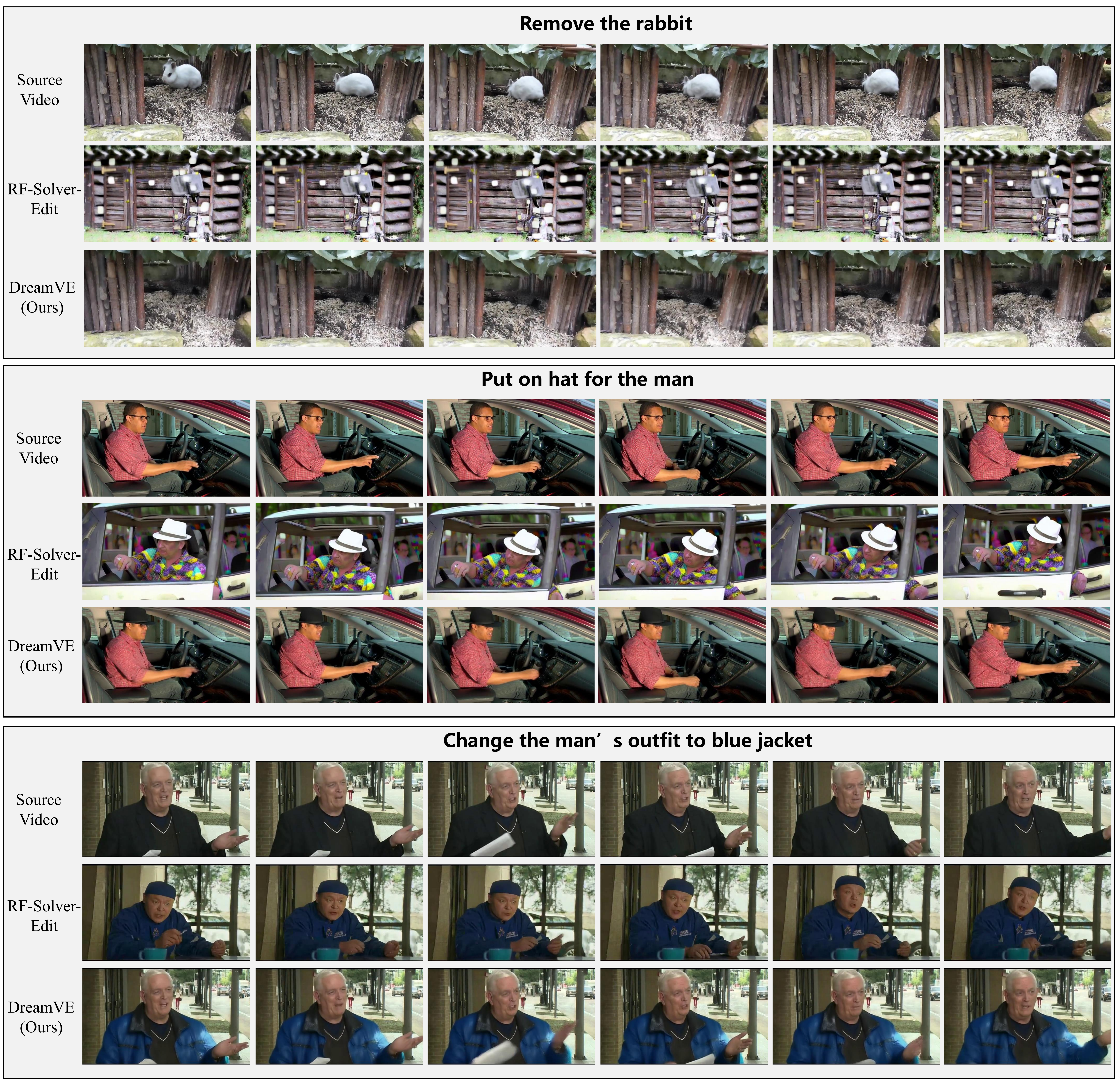}
 }
 \vspace{-2mm}
	\caption{The collage-based synthetic data pipeline (take object replacement as an example).
 } 
	\label{fig:vis2}
\end{figure*}

Here we present a visual comparison with the latest training-free video editing method, RF-Solver-Edit~\cite{rfsolver}. As shown in Figs.\ref{fig:vis1} and\ref{fig:vis2}, our method delivers significantly better editing results than RF-Solver-Edit. Notably, since our approach eliminates the need for the inversion process, DreamVE achieves a $2.2\times$ speedup in editing compared to RF-Solver-Edit (see Tab.~\textbf{3} in the main paper).

\section{ Collage-based Synthetic Data Pipeline}
\label{sec:data}
\begin{figure*}[t]
	\centering
 \resizebox{0.8\linewidth}{!}{
	\includegraphics[height=4cm]{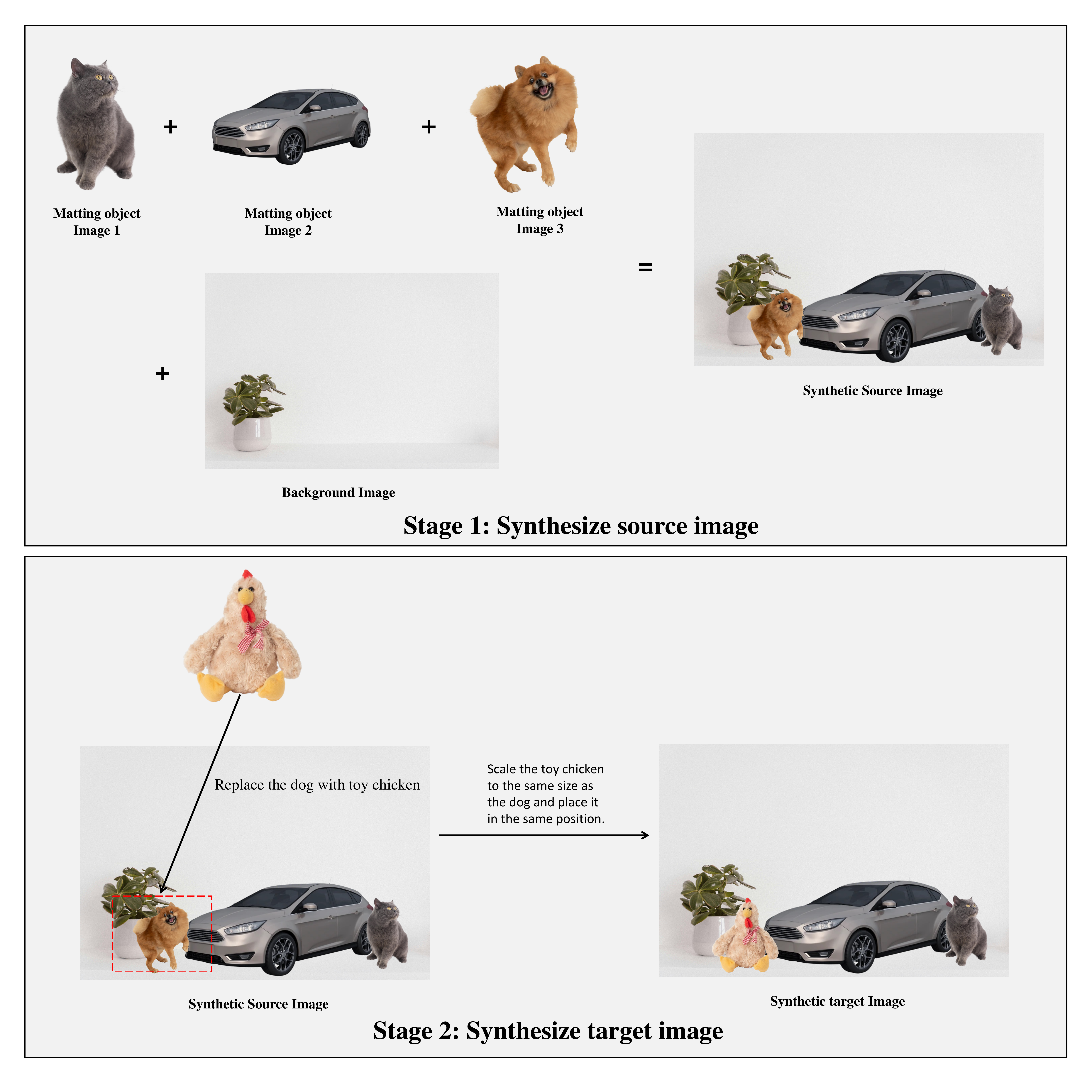}
 }
 \vspace{-2mm}
	\caption{The collage-based synthetic data pipeline (take object replacement as an example).
 } 
	\label{fig:data_pipeline}
\end{figure*}

In this section, we provide further details about our synthetic collage data pipeline introduced in the main paper.
The synthetic data for video editing and image clipping are almost identical, with the main difference being that in the video, both the background and foreground consist of video clips. As shown in Fig.~\ref{fig:data_pipeline}, we select multiple foreground objects from an object material database and paste them onto a background to create the source image. We then perform specific operations on the source image to generate the corresponding instructions and target images. For instance, in the case of object replacement, we select any object from the foreground and replace it with a new one. When placing the new object, it is scaled to match the size of the original and positioned at the same location.

\begin{table*}[h!]\centering

\begin{minipage}{1\columnwidth}\vspace{0mm}    \centering
\begin{tcolorbox} 
    \small
     \hspace{-6mm}

You will receive multiple source image captions describing scenes. Your task is to:
        Modify the both source and target image captions to change the clothing type, accessories, or style of a specific person, animal, or object while keeping the rest of the caption intact.
        Provide the modified caption as the ``source caption'' and ``target caption''. Clearly describe the change you made as an ``instruction''. Include the original and edited clothing types, accessories, or styles as "origin item" and ``edit item''

        Example Input:
        
        Source Caption: ``A child in a coat feeds pigeons amidst bare branches.''

        Example Output:
        \{
        ``source caption'': "A child in a coat feeds pigeons amidst bare branches.",
        ``target caption'': "A child in a T-shirt feeds pigeons amidst bare branches.",
        ``instruction'': "change the child's coat to T-shirt",
        ``orign item'': "coat",
        ``edit item'': "T-shirt"
        \}

        Guidelines:
        Ensure your response strictly follows this JSON format for each caption provided.
        Do not include unnecessary information outside the required fields.
        Focus solely on changes related to facial features or expressions.

\end{tcolorbox}
    
\vspace{-2mm}
\caption{The example of system prompt for GPT-4o to make outfit modification instruction.}
    \label{tab:system_prompt}
\end{minipage}
\end{table*}

\section{ How to Use GPT-4o Make Data?}
\label{sec:gpt4o}
As shown in Tab.~\ref{tab:system_prompt}, we provide an example of how to use the system prompt to enable GPT-4o to generate corresponding editing instructions according to the diverse captions of the real data we provide.

\section{ Video Editing Results Display}
\label{sec:video}
To provide a more comprehensive view of the video editing results presented in our paper, we have included the full mp4 videos in the ``videos'' folder of the supplementary materials. The ``head-demo'' folder contains the editing demo featured at the beginning of the paper, while the ``visual results'' folder includes the comparison results shown in the main paper Fig.~\textbf{6}.

\end{document}